\documentclass[11pt]{article}
\usepackage{bm, amsmath, amsfonts, amssymb,lscape}
\usepackage{graphicx}
\usepackage{amssymb}
\usepackage{bm}                
\usepackage{mathrsfs}      
\usepackage{amsmath, amsthm}
\usepackage{graphicx,epsf,subfigure,pstricks,pst-node,psfrag,amsthm,amssymb,amsmath}
\usepackage{float}
\usepackage{array}
\usepackage{hyperref}
\usepackage{rotating}
\usepackage{natbib}
\usepackage{listings}

\usepackage{hhline}
\usepackage{graphicx}
\usepackage{amssymb}
\usepackage{bm}                
\usepackage{mathrsfs}      
\usepackage{amsmath, amsthm}
\usepackage{graphicx,epsf,subfigure,pstricks,pst-node,psfrag,amsthm,amssymb,amsmath}
\usepackage{float}
\usepackage{array}
\usepackage{hyperref}
\usepackage{authblk}
\usepackage{natbib}

\numberwithin{equation}{section}

\usepackage{JASA_manu}

\begin{document}

\title{A fatal point concept and a low-sensitivity quantitative measure for traffic safety analytics}
\author{Shan Suthaharan}
\affil{Department of Computer Science \\ University of North Carolina at Greensboro \\ Greensboro, NC 27402, USA.}
\date{}
\maketitle

\begin{center}
\textbf{Abstract}
\end{center}
The variability of the clusters generated by clustering techniques in the domain of latitude and longitude variables of fatal crash data are significantly unpredictable. This unpredictability, caused by the randomness of fatal crash incidents, reduces the accuracy of $crash frequency$ (i.e., counts of fatal crashes per cluster) which is used to measure traffic safety in practice. In this paper, a quantitative measure of traffic safety that is not significantly affected by the aforementioned variability is proposed. It introduces a fatal point -- a segment with the highest frequency of fatality -- concept based on cluster characteristics and detects them by imposing rounding errors to the hundredth decimal place of the longitude. The frequencies of the cluster and the cluster's fatal point are combined to construct a low-sensitive quantitative measure of traffic safety for the cluster. The performance of the proposed measure of traffic safety is then studied by varying the parameter $k$ of $k$-means clustering with the expectation that other clustering techniques can be adopted in a similar fashion. The 2015 North Carolina fatal crash dataset of Fatality Analysis Reporting System (FARS) is used to evaluate the proposed fatal point concept and perform experimental analysis to determine the effectiveness of the proposed measure. The empirical study shows that the average traffic safety, measured by the proposed quantitative measure over several clusters, is not significantly affected by the variability, compared to that of the standard crash frequency.



\newpage

\newtheorem{theorem}{Theorem}
\newtheorem{lemma}{Lemma}
\newtheorem{remark}{Remark}
\newtheorem{definition}{Definition}
\newtheorem{example}{Example}
\newtheorem{corollary}{Corollary}
\newtheorem{ex}{}
\newtheorem{proposition}{Proposition}

\section{Introduction}

Fatal crash incidents on highways, cities, and urban/rural roads are significant concerns to commuters and residents of the United States. Although it is a national problem \citep{schultz2011transportation}, this paper only considers the fatal crash incidents in the State of North Carolina; however, it can be easily extended to address the same problem in other states. One of the most recent articles \citep{pour2016analyzing} discusses the fatal crash accidents in North Carolina, but is related to the severity of motorcycle crashes. Fatal crash incidents are random and possibly correlated with the locations \citep{abdel2007crash}, where we can represent the locations using the variables longitude and latitude. As a result, they display interesting patterns in the form of clusters of locations over a very large domain, such as cities and states. These cluster formations are generally observable from the scatter plot of fatal crash incidents with longitude (x-axis) and latitude (y-axis) as location variables. Hence, the statistical pattern recognition approaches \citep{suthaharan2015machine} can be adopted to study this problem efficiently. 

The clusters of locations  can be extracted using clustering techniques, such as $k$-means \citep{macqueen1967some}, k-means++ \citep{arthur2007k}, and k-medoids clustering \citep{jin2011k}. When clusters are extracted, it is expected that many strong clusters will overlap with the busy traffic areas and population dense regions, such as major cities and arterial highways. Hence, the frequency of fatal crash incidents can be observed from the cluster locations using visualization techniques \citep{ramos2015detection}. This crash frequency can be used as a quantitative measure of traffic safety to identify the locations that face traffic safety problems. The crash frequency is a simple measure, and it can be used to represent the count of fatal crash incidents as well: the higher the fatal crash frequency the lower the traffic safety. The authors of \citep{seri2015exploring} performed their study by grouping the crash frequency analysis into Crash Frequency-Based Analysis and Crash Rate-Based Analysis. They defined the crash frequency analysis by the absolute crash frequency with respect to each crash type and its severity, and the crash rate  analysis as the combination of absolute crash frequency and traffic volume at the crash location. Thus, we can group the quantitative measures of traffic safety into absolute and relative fatal crash frequency measures for our study, where the relative fatal crash frequency measure quantifies the fatal crash frequency relative to some other factors that influence the fatal crash incidents.
\begin{figure*}[!t]
\centering
\includegraphics[scale=0.7]{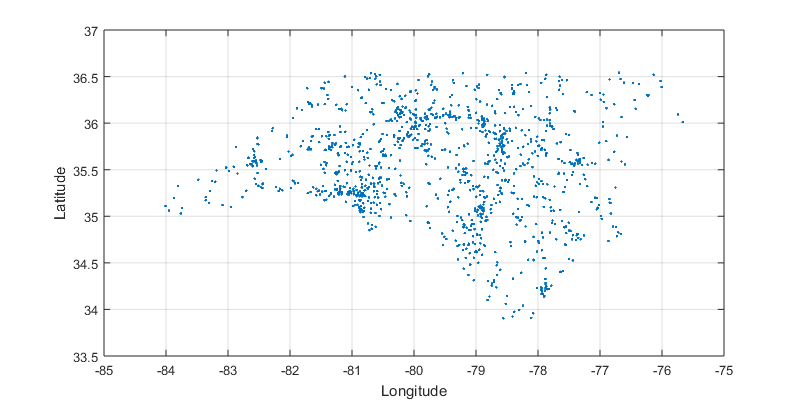}
\caption{The locations information, Longitude and Latitude, extracted from 2015 NC Fatal Crash data set of FARS data source are plotted in this figure. It shows several clusters that overlap with major cities (e.g. Charlotte, Greensboro, and Raleigh) and highways (e.g., I-85 and I-95) of NC.}
\label{fig:visual}
\end{figure*}

As stated in \citep{lovegrove2006macro}, a limited number of quantitative measures with the relative crash frequency can be found in the transportation literature. The paper by \citep{liu2016vehicular} focused on injury crash frequency and severity using absolute crash frequency. Their focus is on the data associated with the crash counts on intersections. The authors of \citep{chiou2013modeling} reported the contributors to crash frequency and severity differ in many cases; hence, they proposed an architecture based on a multinomial generalized Poisson (MGP) model, which allowed them to analyze the frequency and severity of traffic incidents together. The paper \citep{claros2017safety} focused on evaluating red light cameras for traffic safety. They also used absolute crash frequency as a measure to make their conclusions such as ``the reduction in angle crashes due to red light camera'' and ``the increase in rear-end crashes.'' Hence, the crash frequency plays a significant role in crash incidents analysis. 

In the paper \citep{chiou2015modeling}, the authors analyzed what they called multi-period crash frequency and severity by using a Poisson distribution model that they proposed. The model utilized spatial and temporal dependencies of the traffic incidents; hence, it provides a type of relative crash frequency analysis. However, they do not consider the relative frequency with respect to a fatal point that has the highest frequency of fatality. The paper \citep{abdel2007crash} presented a location-based traffic safety analysis using crash frequency as a measure. They studied the traffic safety performance at both the individual crash level and collective crash level, and in both case, the crash frequency played a major role. \citep{ivan2015identification} studied the effect of low-light conditions on accident (crash) frequency in a city in Romania. Their findings show that the traffic accident count (i.e., the crash frequency) is linearly correlated with the low-light condition. \citep{gitelman2010designing} presented a composite safety indicators for traffic safety and determined that the frequency (or rate) of fatal crash incidents is not sufficient to determine the traffic safety performance.

The authors \citep{montella2010comparative} used four quantitative evaluation criteria to compare seven crash-hotspot identification methods for road safety. The main quantitative measures they used were the crash frequencies associated with property damage, crash rate over time, and empirical Bayes estimate of crash frequency. The authors \citep{zhang2012crash} studied the mathematical relationship between crash frequency and the characteristics of road segments associated with urban roadways. They used generalized additive models with crash frequency to study the mathematical relationship. Hence, It is clear from the transportation research literature that the crash frequency is the major influencer in traffic safety research. Therefore, it will be useful to the traffic safety research community if a better quantitative measure of traffic safety is proposed and evaluated. 
\begin{figure*}[!t]
\centering
\includegraphics[scale=.7]{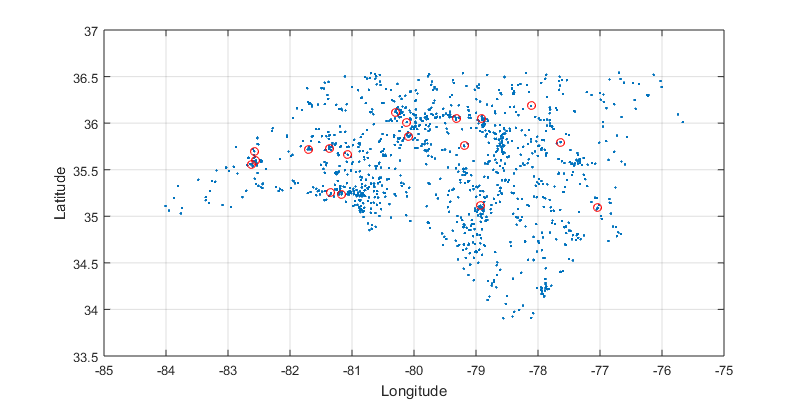}
\caption{It shows the clusters identified by the $k$-means clustering when applied to the data plotted in Figure 1. The highlighted points in red are the centroids (not an actual fatal crash location) of the cluster identified by the $k$-means clustering.}
\label{fig:k-means}
\end{figure*}

\section{Research Motivation}
One of the motivating factors is the limited research on developing a relative quantitative measure for fatal crash incidents. The Transportation Research Board (TRB) provides a database at TRID: \url{https://trid.trb.org/}, where the articles on transportation research can be found. A search on this database indicated the need for such an important measure. Another motivating factor is that the widely used crash frequency measure is very sensitive to the cluster variability -- caused by the randomness of the fatal crash incident -- when the clusters are created by clustering techniques. The third motivating factor is the need for the number of clusters, \textit{a priori}, as a parameter to a clustering technique, while the goal of applying clustering is to find the right number of clusters that can help characterize crash locations. Hence, this paper presents a novel \textit{fatal point concept} and a validation method using k-means clustering and FARS data to reduce the above problems. 

\section{An Example Using FARS Data Source}
The cross-sectional fatal crash data in Fatality Analysis Reporting System (FARS) indicate that the spread of fatality incidents on highways, major cities, and rural areas in the State of North Carolina (NC) is a significant threat to commuters and residents \citep{FARS}. FARS is a data source that provides factual information about fatal crash incidents occurring in the United States. As mentioned earlier, we only considered 2015 NC fatal crash data of FARS to study the patterns of fatal crash incidents in NC. We can clearly observe in Fig. \ref{fig:visual} that the location information (longitude and latitude) of fatal crash incidents in the FARS data sets draws the map of North Carolina. The evolution of such a state map from the fatal crash incidents is an indicator that the traffic safety is a serious problem in NC. 

Therefore, it is important to study this fatal crash data set by utilizing suitable data mining techniques to characterize fatal crash locations, quantify traffic safety, and develop solutions to the spatially oriented traffic safety problem in NC. The most commonly used quantitative measures in traffic safety analytics is crash frequency; therefore, we used it in this example along with $k$-means clustering to describe these data sets. For example, the application of $k$-means clustering with, $k$=18 to the fatal crash data in Fig. \ref{fig:visual}, produced the 18 clusters shown in Fig. \ref{fig:k-means}, which highlights some of the meaningful clusters. Also note that the highlighted points in red are the centroids of the clusters, not the actual locations where fatal crash incidents occurred. The scatter plots in these figures demonstrate that the frequency of fatality is very high in major cities, such as Charlotte, Greensboro, and Raleigh, and low in the northeastern areas of NC. It also shows that the frequency of crashes are high along the main arterial highways, such as I-40, I-85, and I-95. This distribution of clusters would naturally encourage one to apply a clustering technique repeatedly with increasing values of the parameter $k$ to detect more clusters as needed. 

The repeated application of a clustering method creates variability in the clusters, which is an effect of the randomness of the fatal crash incidents. This problem can be conceptualized with a simple example: if we consider two clusters with the same crash frequencies, one may interpret that those two clusters have the same traffic safety quantitatively. However, if one of these clusters has a fatal point that has the highest fatal crash incidents, and the other cluster has several isolated fatal crash incidents, then the assumption of the same traffic safety for these two clusters may not be appropriate (i.e., the variability of the clusters can affect the clusters' crash frequencies). This analogy is still valid when we have two clusters with two different crash frequencies in which one cluster has smaller segments of locations with very high fatal crash incidents. The authors \citep{claros2017safety} also stated that the crash frequency (i.e., an absolute crash frequency) may not be suitable to characterize crash incidents or quantify traffic safety. Thus, it is applicable to the characterization and quantization of fatal crash incidents. 

The FARS data set is freely available, and its accessibility is not that limited; hence, it supports the reproducible research without any restrictions. It consists of several data sets in spreadsheets, which makes a significant convenience to many applications. We have used the accident related 2015 NC fatal crash data, which includes 1,275 fatal crash incidents. However, we had to perform a minor cleansing to identify and remove the outliers with unusual values, such as 888.8888 and 999.9999. We found such outliers in 12 longitude or latitude variables; hence, the removal of these records reduced the number of records to 1,263, which is still large enough to perform the analysis. The number of independent variables found in this data is 51, but we are interested only in longitude and latitude variables associated with each fatal crash incident in this paper.

\section{Proposed Approaches}
The proposed approaches bring two new contributions to the traffic safety research domain. The first approach is the concept that leads to a development of a ratio as a quantitative measure to quantify traffic safety. The second approach is the fatal point detection with the induction of rounding errors, where the fatal point crash frequency is a major contributor to make the proposed quantitative measure less sensitive to cluster variability.

\subsection{Proposed Fatal Point Concept}
\begin{figure}[t]
\centering
\includegraphics[scale=.8]{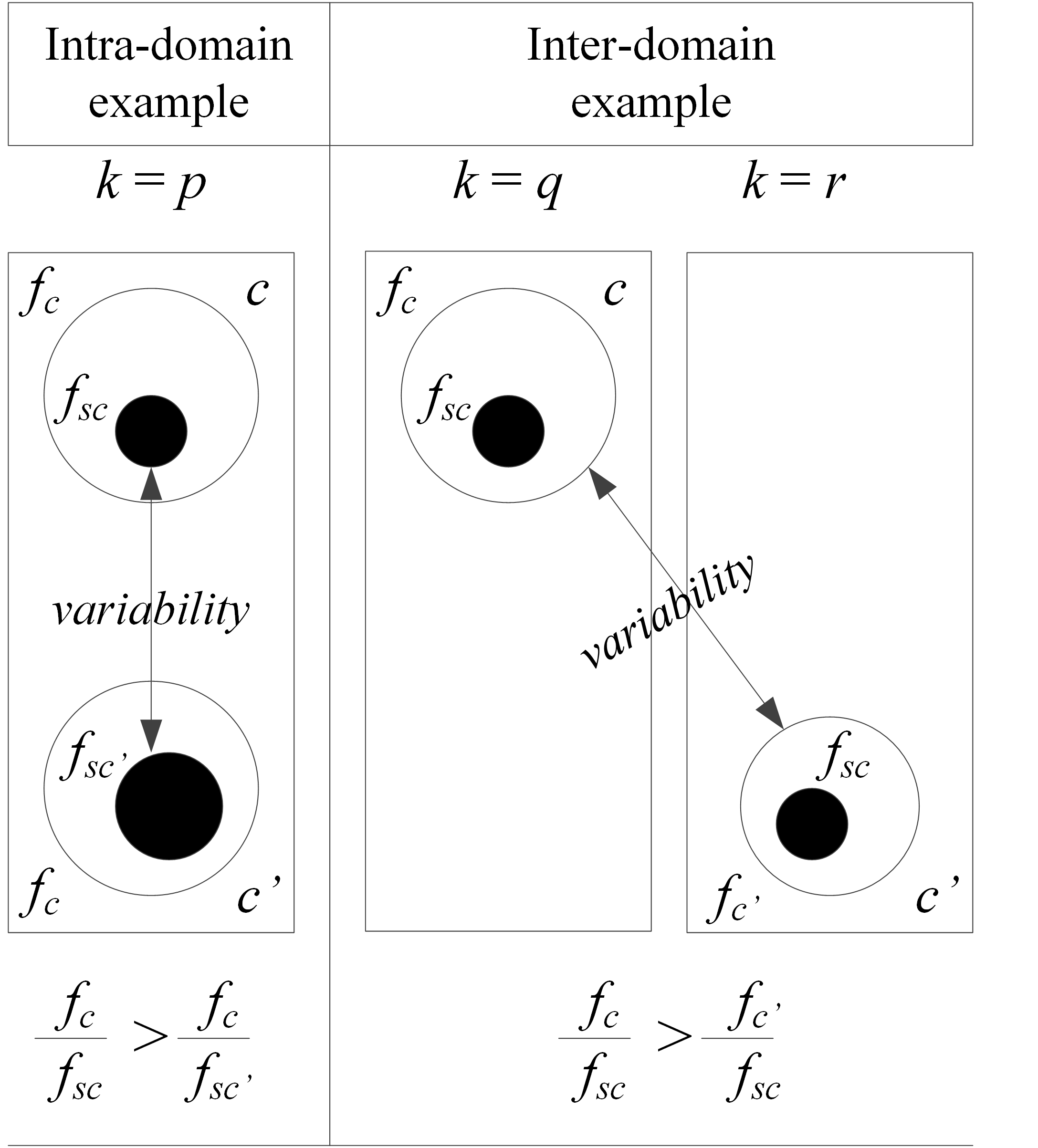}
\caption{There are many possible cases of variabilities, but the two most important ones are illustrated in this figure. In both cases the cluster $c'$ has less traffic safety rating than the rating of the cluster $c$.}
\label{fig:concept}
\end{figure}

The proposed concept is presented in Fig. \ref{fig:concept}, where the size of the shapes represents the proportionality of the frequency rather than the size of the cluster region on the map of fatal crash incidents. To explain the concept, let us first define a set of variables, and a concept: the variable $f_c$ that represents the fatal crash frequency of a cluster $c$, and the variable $f_{sc}$ that represents the frequency of the \textit{fatal point} of a cluster $c$. The fatal point is a new concept, and it is defined as follows:
\begin{itemize}
\item{The \textit{fatal point} is a logical location that is mapped to a set of locations (longitude and latitude) that form a segment with the highest frequency of fatality within a cluster.}
\end{itemize}
Fig. \ref{fig:concept} presents two examples that justify the reason for using a fatal point in the development of the proposed quantitative measure. The intra-domain example shows the variability between two clusters in a domain of clusters generated using a single value of $k$ in $k$-means clustering. Similarly, the inter-domain example shows the cluster variability between two clusters from two domains of clusters generated using two different values of $k$ in $k$-means clustering. 

The intra-domain example shows two clusters, $c$ and $c'$, that have the same crash frequency $f_c$. It also shows their fatal point frequencies of fatality, $f_{sc}$ and $f_{sc'}$, where $f_{sc'} > f_{sc}$ (one of the variabilities mentioned earlier). Hence, the ratios $f_c/f_{sc}$ and $f_c/f_{sc'}$ satisfy the following inequality:
\begin{equation}
\frac{f_c}{f_{sc}} > \frac{f_c}{f_{sc'}}.
\label{eq:ratio1}
\end{equation}
It shows that cluster $c'$ has lower traffic safety than cluster $c$. Similarly, the inter-domain example shows that clusters $c$ and $c'$ have distinct crash frequencies $f_c$ and $f_{c'}$, where $f_{c} > f_{c'}$  (the second variability), but their fatal points have the same frequencies of fatality $f_{sc}$. Hence, the ratios $f_c/f_{sc}$ and $f_{c'}/f_{sc}$ satisfy  the following inequality:
\begin{equation}
\frac{f_c}{f_{sc}} > \frac{f_{c'}}{f_{sc}}.
\label{eq:ratio1}
\end{equation}
It also shows the cluster $c'$ has a lower traffic safety than cluster $c$. Different examples can be generated using this analogy (or patterns), but these two patterns can form the basis for them.

\begin{figure*}[!t]
\centering
\includegraphics[scale=.65]{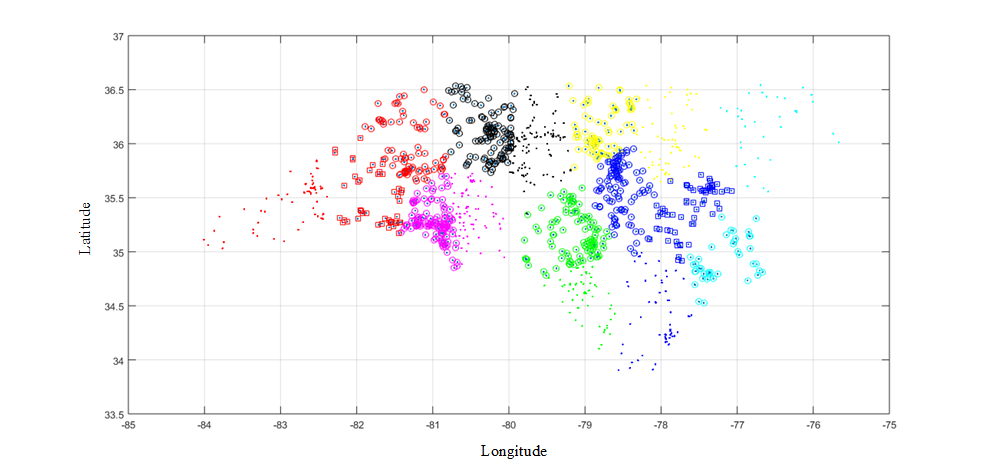}
\caption{The spatial variability of the clusters are are shown in different colors and markers. It shows the 16 clusters of the k-means clustering, with $k$=16.}
\label{fig:cluster}
\end{figure*}

\subsection{Proposed Quantitative Measure}
The proposed measure is devised based on the aforementioned concept and the ratio between the cluster frequency and the fatal point frequency. Hence, we represent the quantitative measure of traffic safety using the variable $u_c$, which varies with respect to the variability of cluster $c$, and define it as follows:
\begin{equation}
u_c = \frac{f_c}{f_{sc}},
\label{eq:qmts}
\end{equation}
where $f_{sc}$ is the frequency of the segment that has the higher frequency of fatality (i.e., the fatal point). The idea is to make the new quantitative measure $u_c$ uncorrelated to the cluster frequency $f_c$ so that the new measure is less sensitive to the variability of clusters, which is generated by the randomness of fatal crash incidents, than the cluster frequency $f_c$, which is generally affected significantly by cluster variability. 

\subsection{Proposed Fatal Point Detection}
A method is proposed to define and detect the fatal point of a cluster by dividing the cluster into vertical segments of about 0.7 miles width along the longitude variable. The reason for the selection vertical segments is that our observation of horizontal traffic flow pattern over the state of NC; however, the future research will include the horizontal segments as well. In this paper, we also have secured the width of the segment to 0.7 miles and it will be optimized in the future research using an empirical approach. Each segment can be interpreted as a set of fatal crash locations in the proposed approach. These sets are created by rounding the longitude values of a cluster to their hundredth decimal places based on the information available \citep{SE}. The rounded longitude locations are called the logical locations, and they create redundancy in the fatal crash locations, thus forming the aforementioned sets. The redundancies are used, along with the statistical measure ``mode,'' to obtain frequencies of fatality for these sets. The set with the highest frequency of fatality of the cluster is defined as the fatal point and is detected by the approach.


\section{Results and Performance Evaluation}
The number of clusters that we considered for the $k$-means clustering varies from $k$=8 to $k$=128, with an increment of 1. Hence, they produce 121 domains of clusters. The cluster domain with $k$=16 is selected to explain the steps of the simulation. When $k$-means clustering with $k$ = 16 is applied to the two-dimensional map (longitude and latitude) of 2015 NC fatal crash data in Fig. \ref{fig:visual}, it produced the spatial characteristics of the clusters presented in Fig. \ref{fig:cluster} and were highlighted using different colors and markers. It clearly characterizes the fatal crash incidents as clusters of locations that are meaningful to describe the map of North Carolina. The fatal crash frequencies $f_c$, where $c = 1, 2, \dots, 16$, of the 16 clusters are calculated and presented in the second column of Table 1. Also note that the first column of the table provides the labels for the 16 clusters. 

We can observe from the table that the largest value of the variable $f_c$ is 141, indicating the cluster $c_9$, has the largest number of fatal crashes in NC in 2015, which is associated with the areas (Charlotte) highlighted in “magenta” color. Similarly, the smallest value of $f_c$ is 37, indicating the cluster $c_6$, has the smallest number of fatal crashes in NC in 2015, which is associated with the areas (northeast of NC) in “cyan” color. So, using the crash frequency as a measure, we can say that cluster $c_9$ has the lowest traffic safety and cluster $c_6$ has the highest traffic safety among the 16 clusters determined by $k$-means clustering. Hence, it is important to note that the larger crash frequency means lower traffic safety (i.e., the crash frequency provides an opposite measure for traffic safety). In the third column of the Table 1, the normalized values $N(f_c)$ of $f_c$ are also presented to transform the measure to a common scale.

\begin{figure*}[!t]
\center
\includegraphics[scale=.78]{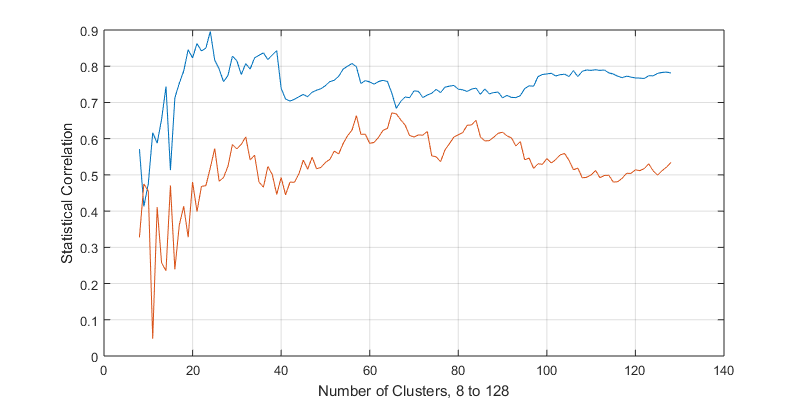}
\caption{The correlations between $f_{sc}$ and $f_c$ in blue and $u_c$ and $f_c$ in red are presented.}
\label{fig:corr}
\end{figure*}

\subsection{Results from Fatal Point Detection}
The fatal crash frequency $f_c$ provides the information about the number of fatal crash locations with longitude and latitude values. For example, the number of fatal crash locations in cluster 1 is 74; therefore, this cluster has at most 74 distinct longitude values and their corresponding latitude values. In the fatal point detection, we are interested in the vertical segments; therefore, we induced rounding errors to longitude values by rounding them to their hundredth decimal place. Then the statistical measure ``mode'' is used to detect the number of fatal crash locations in close proximity, which we call logical fatal crash locations. The set of locations associated with the most occurring logical fatal crash location is recorded, and we call it a fatal point. The number of fatal crash incidents in the fatal points of the 16 clusters are listed in the fourth column of Table 1. One location from each fatal point of the clusters are also listed in the fifth column of the table to help the readers associate them visually with the clusters highlighted in Fig. \ref{fig:cluster}. For this particular domain of clusters (i.e., $k$=16), $f_{sc}$ and $f_c$ are correlated with the correlation value of 0.7129. It does not mean the high correlation is guaranteed for other values of $k$. 

\begin{table}[h]
\caption{A set of statistical information of fatal crash incidents observed at the 16 clusters that are identified by the k-means clustering is presented.}
\vspace{3mm}
\centering
\begin{tabular}{|r|r|r|r|c|r|}
\hline
$\bm{c}$ & $\bm{f_c}$ & $\bm{N(f_c)}$ & $\bm{f_{sc}}$ & $\bm{(g, t)}$ of $\bm{s_c}$ & $\bm{N(u_c)}$\\
\hhline{|=|=|=|=|=|=|}
1 & 74 & 0.3558 & 7 & (-82.6035, 35.5930) & 0.1556\\
\hline
2 & 63 & 0.2500 & 4 & (-78.1590, 34.5339) & 0.4330\\
\hline
3 & 121 & 0.8077 & 8 & (-79.9919, 36.0909) & 0.3996\\
\hline
4 & 80 & 0.4135 & 6 & (-81.3178, 35.7414) & 0.3036\\
\hline
5 & 105 & 0.6538 & 9 & (-79.8266, 36.0604) & 0.2143\\
\hline
6 & 37 & 0.0000 & 3  & (-76.2385, 36.3045) &  0.2500\\
\hline
7 & 70 & 0.3173 & 5 & (-80.6414, 35.3514) & 0.3393\\
\hline
8 & 62 & 0.2404 & 4 & (-78.9951, 34.6181) & 0.4196\\
\hline
9 & 141 & 1.0000 & 11 & (-81.3413, 35.2512) & 0.2760\\
\hline
10 & 79 & 0.4038 & 3 & (-78.7731, 35.9044) & 1.0000\\
\hline
11 & 103 & 0.6346 & 4  & (-79.3048, 34.9407) & 0.9688\\
\hline
12 & 67 & 0.2885 & 3 & (-77.6371, 35.7966) & 0.7857\\
\hline
13 & 47 & 0.0962 & 4 & (-77.5481, 34.8875) & 0.2188\\
\hline
14 & 99 & 0.5962 & 5 & (-78.6750, 35.8333) & 0.6550\\
\hline
15 & 46 & 0.0865 & 6 & (-81.5502, 35.2716) & 0.0000\\
\hline
16 & 69 & 0.3077 & 6  & (-77.0761, 35.5655) & 0.2054\\
\hline
\end{tabular}
\end{table}

Therefore, we repeated the correlation analysis with the values of $k$ from 8 to 128, and the results are presented in Fig. \ref{fig:corr} in blue. It shows the following: when the number of clusters determined by k-means clustering is low (i.e., the size of the cluster is large), the correlations between $f_c$ and $f_{sc}$ are also low (about 0.4). When the number of clusters are increased, the correlations $f_c$ and $f_{sc}$ are also increased and stabilized at about 0.75. Hence, they all make sense that when the cluster sizes become smaller, the cluster and the fatal point become closer. With this result, we can validate the use of $k$-means clustering and the induction of rounding errors. Although, this research considered the k-means clustering only, the Density-based spatial clustering of applications with noise \citep{ester1996density}, Ordering points to identify the clustering structure \citep{ankerst1999optics}, k-means++, and k-medoids will be used in the future research to evaluate the concept of fatal point.   

\subsection{Results from Quantitative Measure}
The traffic safety measure ($u_c$) in equation (\ref{eq:qmts}) is then calculated and normalized to $N(u_c)$. The results are presented in the sixth column of Table 1. The correlations between $N(u_c)$ and $N(f_c)$, with the number of clusters varying from 8 to 128, are calculated and presented in Fig. \ref{fig:corr} in red.  It shows the following: initially when the number of clusters determined by $k$-means clustering is low, the correlation between $u_c$ and $f_c$ are also low (about 0.05). Later when the number of clusters are increased, the correlations $u_c$ and $f_c$ are also increased and stabilized at about 0.5. Now, comparing the correlation results with blue and red, we can determine that the proposed quantitative measure, as a random variable, has a lower correlation with the cluster frequency $f_c$ than the fatal point frequency $f_{sc}$. We are also interested to know whether these correlations are correlated or not correlated themselves with respect to the number of clusters. Hence, we calculated the correlation of these correlations and obtained a very low correlation value of 0.1452. This indicates, when $f_c$ and $f_{sc}$ are correlated, it is less likely $f_c$ and $u_c$ will be correlated. Similarly, when $f_c$ and $f_{sc}$ are uncorrelated it is less likely $f_c$ and $u_c$ will be uncorrelated. Hence, the selection of $u_c$ as an alternative measure to $f_c$ is appropriate so that we can limit the effect of the cluster variability.

\begin{figure*}[!t]
\center
\includegraphics[scale=.8]{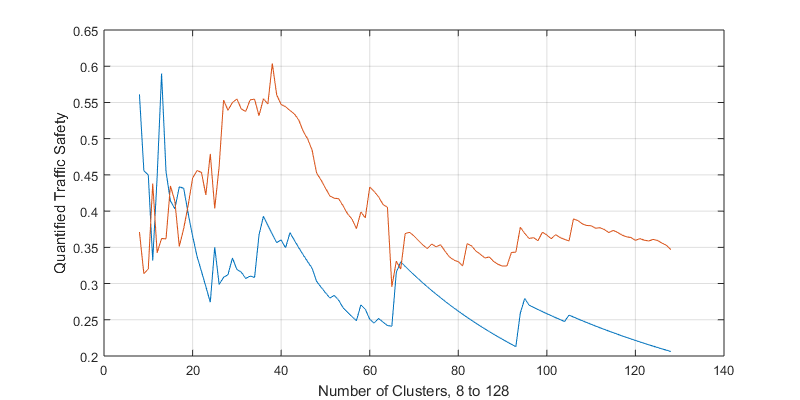}
\caption{The averages of the two measures over 121 domains of clusters are presented.}
\label{fig:mean}
\end{figure*}

\subsection{Performance Evaluation}
\begin{figure*}[!t]
\center
\includegraphics[scale=.7]{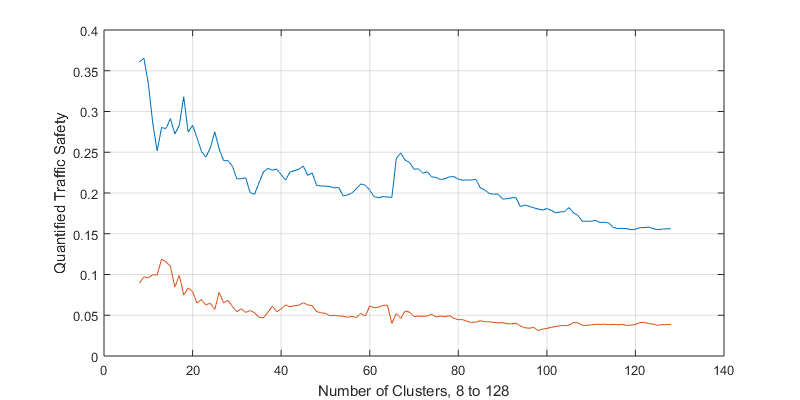}
\caption{The variances of the two measures over 121 domains of clusters are presented.}
\label{fig:variance}
\end{figure*}

The expectation of the proposed quantitative measure is to perform well in terms of measuring traffic safety accurately under the variability issues resulted from the randomness of the fatal crash incidents. Therefore the goal of the performance evaluation is to use the results obtained in the previous subsections and compare them with the results of the standard crash frequency measure. The average traffic safety values of NC per domain are calculated for the normalized traffic safety values ($N(f_c)$ and $N(u_c)$) of the clusters in that domain. As mentioned earlier, there are 121 domains (based on the values of $k$ ranging from 8 to 128); therefore, we have 121 average traffic safety values for NC. These values are presented in Fig. \ref{fig:mean}, where blue represents the averages of $N(f_c)$ and red represents the averages of $N(u_c)$. At the same time, the averages of the variances are also calculated and presented in Fig. \ref{fig:variance}, respectively.

To evaluate the performance of the proposed measure against the standard crash frequency measure, we divided the 121 domains into four groups and analyzed the results in Fig. \ref{fig:mean} and Fig. \ref{fig:variance} together. The parameter $k$ values of 8 to 24 is group 1, 25 to 40 is group 2, 41 to 64 is group 3, and 65 to 128 is group 4 are selected based on the changes that we can observe. Within group 1, both the proposed measure and the standard crash frequency measure show an increase in traffic safety. Note a decrease in blue graph (i.e., decrease in fatal crashes) means an increase in traffic safety. Although both measures increase the traffic safety, we can clearly observe that the variability of the results from the proposed measure is significantly low. This can also be confirmed from the variance graph in Fig. \ref{fig:variance}.

Within group 2, some stability in the traffic safety rating is displayed by the measures; however, the proposed measure has a strong stability value of 0.54, approximately. The crash frequency measure has a very high oscillation, indicating high variability. Interestingly, both of these measures show conflicting results within the next group. That is, the proposed measure indicates the average traffic safety of the state is decreasing, whereas the crash frequency measure indicates it is increasing. The question is, which one is correct? Considering the results of all four groups, the third of the four groups suggest that the proposed measure is less sensitivity to the variability; hence, we can assume that traffic safety is decreasing within that group by accepting the output of the proposed measure by voting. 

The results of group four are also very interesting and useful because the larger $k$ value means the map of NC is divided into a larger number of clusters with smaller cluster sizes. Hence, the results of this group are highly suitable to describe the overall safety of the entire state. The results in Fig. \ref{fig:mean} and Fig. \ref{fig:variance} indicate that the proposed measure is less sensitive to the cluster variability with the convergence to an average traffic safety value of 0.35, along with the variance that is less than 0.05 and the variability that is very small. Hence, using the proposed measure, we may be able to say that the traffic safety in NC is about 35\%. In other words, the ratio between the cluster crash frequency and the fatal point crash frequency is 1 is to 3, indicating high fatal crashes. 

\section{Conclusion}
The proposed measure is a better and alternative measure to the standard crash frequency measure because it is less sensitive to the variability of clusters generated by $k$-means clustering technique. It also reports that the traffic safety rating of NC is 35\% -- indicating high number of fatal crashes -- using the fatal crash location information available in the 2015 NC fatal crash data set of FARS data source. However, it is important to note that the results and findings that are reported in this paper is limited, and further significant research is required to support the findings. Hence, the future research will include the experimental analysis using other location-based clustering algorithms such as the Density-based spatial clustering of applications with noise, ordering points to identify the clustering structure, k-means++, and k-medoids. 

The future research will also include the analyze of the FARS data from other years so that the accuracy of the above rating can be confirmed. Once that research is completed, the FARS data from other states will be utilized to compare the traffic safety ratings between the states. In addition, the effect of the varying width size of fatal point cluster segments, in contrast to the fix width of 0.7 miles, will be studied to determine the optimal size for the fatal point segments. In the present form, the vertical segments are considered due to the horizontal flow of traffic patterns over the state map, and further study will be conducted with the inclusion of horizontal segments. 

\bibliographystyle{model2-names}
\bibliography{refs}

\end{document}